%% file: main.tex
\documentclass[lettersize,journal]{IEEEtran}
\usepackage{amsmath,amsfonts}
\usepackage{algorithmic}
\usepackage{algorithm}
\usepackage{array}
\usepackage[caption=false,font=normalsize,labelfont=sf,textfont=sf]{subfig}
\usepackage{textcomp}
\usepackage{stfloats}
\usepackage{url}
\usepackage{verbatim}
\usepackage{graphicx}
\usepackage{subcaption}
\usepackage{multirow}
\usepackage{amssymb}
\usepackage{adjustbox}
\usepackage{xcolor}

\usepackage{svg}

\begin{document}

\title{Temporal Enhanced Floating Car Observers}

\author{Jeremias Gerner$^{1}$, Klaus Bogenberger$^{2}$, Stefanie Schmidtner$^{3}$
\thanks{$^{1}$Jeremias Gerner is with the Department of AImotion Bavaria, Technische Hochschule Ingolstadt, 85049 Ingolstadt, Germany
        {\tt\small jeremias.gerner@thi.de}}
\thanks{$^{2}$Klaus Bogenberger is with the School of Engineering and Design, Technical University of Munich, 80333 München, Germany
        {\tt\small klaus.bogenberger@tum.de}}
\thanks{$^{3}$Stefanie Schmidtner is with the Department of AImotion Bavaria, Technische Hochschule Ingolstadt, 85049 Ingolstadt, Germany
        {\tt\small stefanie.schmidtner@thi.de}}

}
\maketitle

\begin{abstract}
Floating Car Observers (FCOs) are an innovative method to collect traffic data by deploying sensor-equipped vehicles to detect and locate other vehicles. We demonstrate that even a small penetration rate of FCOs can identify a significant amount of vehicles at a given intersection. This is achieved through the emulation of detection within a microscopic traffic simulation. Additionally, leveraging data from previous moments can enhance the detection of vehicles in the current frame. Our findings indicate that, with a 20-second observation window, it is possible to recover up to 20\% of vehicles that are not visible by FCOs in the current timestep. To exploit this, we developed a data-driven strategy, utilizing sequences of Bird's Eye View (BEV) representations of detected vehicles and deep learning models. This approach aims to bring currently undetected vehicles into view in the present moment, enhancing the currently detected vehicles. Results of different spatiotemporal architectures show that up to 41\% of the vehicles can be recovered into the current timestep at their current position. This enhancement enriches the information initially available by the FCO, allowing an improved estimation of traffic states and metrics         (e.g. density and queue length) for improved implementation of traffic management strategies.
\end{abstract}

\section{Introduction}
Retrieving accurate and detailed traffic state estimations is crucial for implementing effective traffic management strategies to enable smoother, safer, and environmentally friendly transportation. To this day, the most prominent traffic monitoring systems include induction loop sensors built into the pavement and video detectors that are permanently installed within one location. Those stationary observer systems create high costs due to the number of systems needed within the traffic network, including the installation cost of supporting structures and maintenance. Especially induction loops further only gather limited traffic information~\cite{bernas2018survey}. 
\newline
With the advent of Advanced Driver Assistance Systems (ADAS) and Autonomous Driving (AD), an increasing number of vehicles are equipped with various sensors, including cameras, lidar, and radar. These sensors possess the ability to detect and locate other traffic participants within the driving environment~\cite{wu2023virtual, 10373157, meyer2021graph}. As a result, multiple vehicles can share this information via Vehicle-to-Everything (V2X) communication to a central entity, which can gather and analyze the traffic state information. Consequently, the vehicles act as Floating Car Observers (FCOs), which can potentially reduce the need for expensive and low-quality stationary observers. 
\newline
Research into Floating Car Observers (FCOs) has its origins in the Moving Observer (MO) method, first introduced by~\cite{wardrop1954method}. Initially, human observers gauged macroscopic traffic flow characteristics by monitoring vehicular movements in both their own and the opposing direction. Subsequent studies have extended this methodology, leveraging built-in vehicular sensors to estimate traffic states and parameters~\cite{florin2022real, van2018estimating, yunfei2023avaas}. These advancements aim to accurately determine key macroscopic traffic parameters across urban and non-urban traffic environments, including flow, density, and average speed. Given that these analyses are conducted through simulations, the foundation of their calculations is based on estimating the vehicles that can be detected by the FCOs. However, those works relied on simplified assumptions such as simple distance-based methods. An improvement to these methods was presented in~\cite{gerner2023sumo} by introducing camera-based detection emulation within a microscopic simulation. In this work, we utilize this emulation to analyze the potential of FCOs for traffic state gathering, given different penetration rates of FCOs at a signalized intersection within a microscopic simulation.
Limiting our observation to only the currently visible vehicles overlooks the valuable information that can be derived from previously detected vehicles. To leverage their data in the present time step, it is essential to reintegrate them into the current timeframe. Consequently, we explore the prospect of harnessing detections from prior time steps to construct the traffic state in the current period. Finally, we present a data-driven methodology for reintegrating previously detected traffic participants into the current time step.

\section{Floating Car Observer Based Traffic Information}\label{sec: FCO}
\begin{figure}[tb]
    \begin{minipage}[t]{0.7\columnwidth}
        \includegraphics[width=0.98\textwidth, trim={25mm 20mm 0 0}, clip]{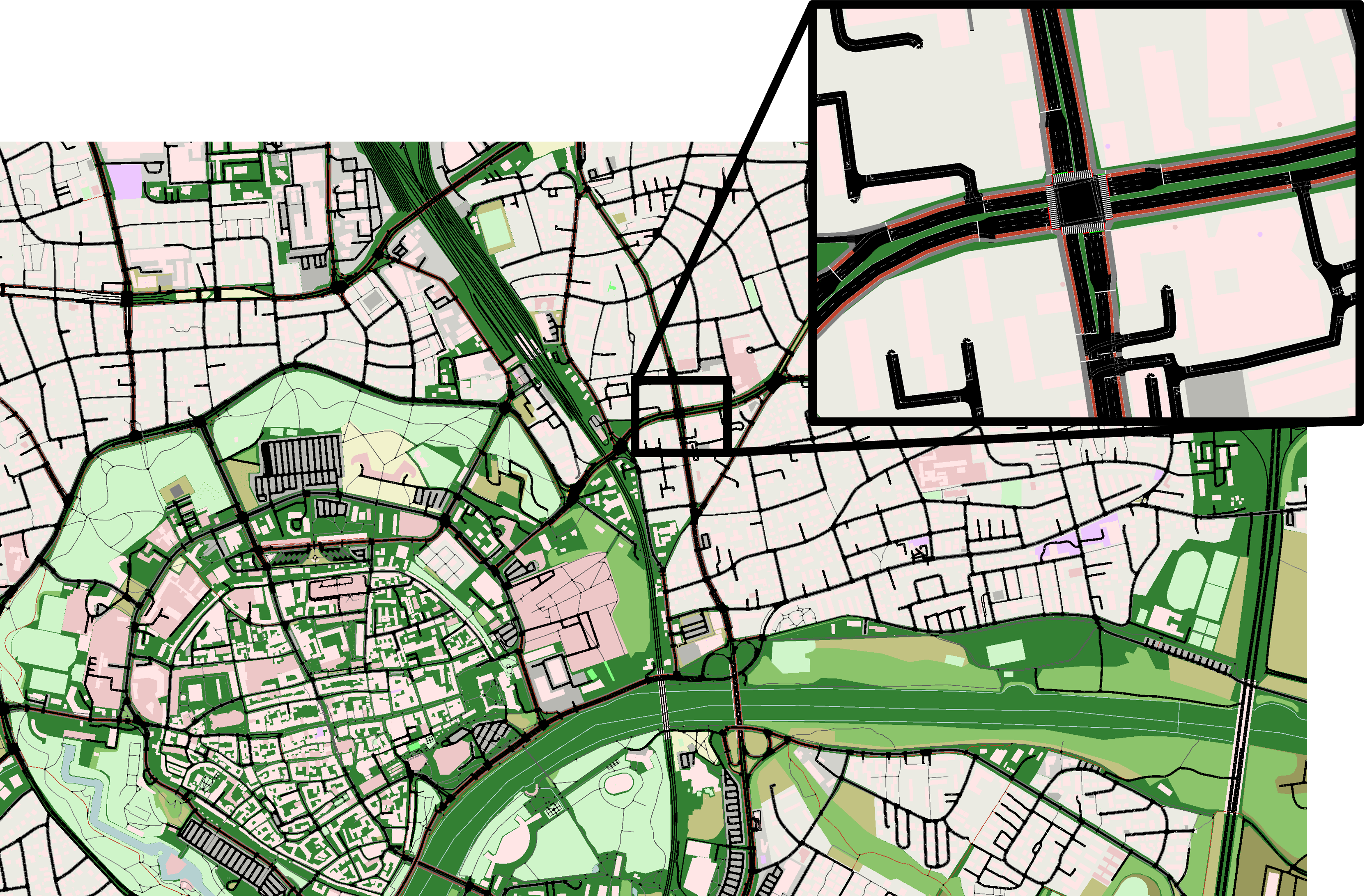}
    \end{minipage}%
   \begin{minipage}[t]{0.3\columnwidth}
      \includegraphics[width=0.9\textwidth]{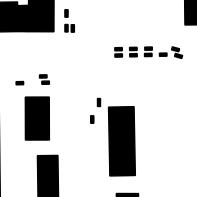}
  \end{minipage}
    \caption{SUMO digital twin of the traffic network of Ingolstadt created by~\cite{harth2021automated} with a highlight set for the intersection investigated in this work (left) and an example BEV representation as input for the BEV detection approach (right).}
    \label{fig: intersection}
\end{figure}

FCOs have the potential to replace stationary sensors, providing a cheaper and potentially richer representation of the traffic state. However, given a certain penetration rate, not all traffic participants can be detected at any given time. Some traffic participants may be out of the sensors' range or occluded, leaving them unavailable to generate a traffic state representation. Hence, before implementing FCOs in the real world, it is essential to elaborate on their effectiveness in a simulation, favorably in microscopic simulations, which are advantageous for their rapid configuration time and simulation speed~\cite{gerner2023sumo}. Our work utilizes the microscopic Simulation of Urban MObility (SUMO)~\cite{behrisch2011sumo}. Specifically, we use the SUMO simulation introduced in~\cite{harth2021automated}, which represents a digital twin of the traffic network of Ingolstadt. Additionally, it provides a realistic traffic demand. Using the simulation, we analyze the effectiveness of FCOs for an individual intersection within the traffic network of Ingolstadt, which is shown in Figure~\ref{fig: intersection}. We especially investigate traffic demands between 6~AM to 9~AM, a timeframe encompassing both high and low traffic volume, representing the intersection's daily traffic flux. This will give insights into the possibility of using the resulting traffic state representation for traffic light control algorithms. 

\subsection{Emulating Floating Car Observers in Microscopic Simulation}\label{subsec: sumo_FCO}
To analyze the potential of FCOs in gathering traffic state information, a realistic 3D detection of traffic participants within a microscopic simulation is necessary. We utilize the camera-based emulation methods introduced in~\cite{gerner2023sumo}. Here, a Computer Vision (CV) approach was introduced, which can approximate the detectability of traffic participants in the view of an FCO using up to four camera sensors and the KITTI dataset~\cite{Geiger_KITTI_2013} as a reference. An alternative approach known as the Bird's Eye View (BEV) method has been introduced in~\cite{gerner2023sumo} to address the issue of slow speed associated with the CV method. This technique enables the direct inference of the detectability of traffic participants by leveraging a two-dimensional BEV representation of the traffic situation centered around the current FCO. Artificial neural networks are used to train the system for detecting traffic participants. The system takes vectors from the FCO to the traffic participant to detect and BEV representation as input, along with the binary detectability labels from the CV approach as ground truth. 

\input{images/multi_images/detected_sequence}
Investigating a single intersection in this work, we created a dataset tailored to this intersection. We curated this dataset to train the BEV approach with high accuracy. Specifically, we only include data in the dataset if the FCO is within a 100-meter radius of the intersection. Unlike the initial work, which uses a snip of the SUMO GUI, we have generated a distinct representation for the BEV that includes traffic participants and buildings. An example is shown in Figure~\ref{fig: intersection}. This allows a simplified representation of a single color channel, resulting in more compact network architectures. Moreover, separating the BEV representation generation from the SUMO GUI leads to faster generation of the BEV representation, further increasing the inference speed of the BEV approach. Upon completing the training phase, the convolution-based model attained high accuracy, specifically 93.9\% in the four-camera setting, when tested on previously unseen scenarios. I.e., the BEV approach approximates the detections of the CV approach with an accuracy of 93.9\% while only requiring 0.2 seconds to evaluate the scenario of one FCO. Consequently, this work will proceed with the BEV approach for estimating the 3D detectability in subsequent investigations.
\subsection{Analysis of Vehicle Detectability by FCOs}\label{subsec: FCO}
Utilizing the trained model for the BEV approach to determine which vehicle would be detectable by an FCO in a comparable real-world scenario, it can now be analyzed and investigated which vehicles are available for generating a traffic state representation at different timesteps within the simulation. Thus, Figure~\ref{fig: seen_sequence} shows a sequence of time steps and highlights which vehicles are recognized, which are not recognized, and which are FCOs themselves at the respective time step. Hence, let $V_t$ be the set of vehicles that are within the radius $r$ around the investigated intersection at the current time step $t$. We express $V_{d,t} \subset V_t$ to represent the subset vehicles detected or acting as FCOs. If the simulation is carried out for a longer period, a distribution can be built from the data that shows the relation from $V_{d,t}$ to $V_t$. I.e. it shows the relative detectability of vehicles that can be used for generating a traffic state representation. Figure~\ref{fig: seen_distribution} visualizes this distribution for different penetration rates of FCOs. 

\subsection{Analysis of Temporal Enhancing Potential}\label{subsec: temp_FCO}
\begin{figure}[t]
    \centering
    \includegraphics[width=.96\columnwidth]{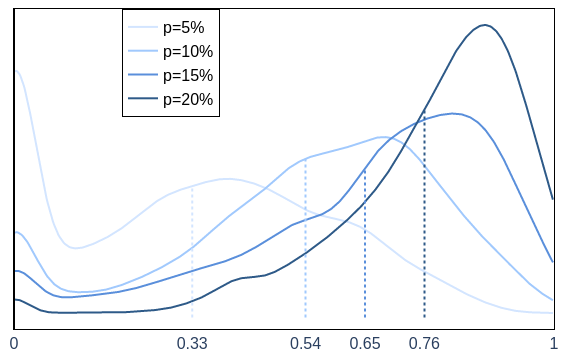}
    \caption{Distribution of $V_{d,t}$ relative to $V_t$ at varying penetration rates of FCOs across the investigated intersection over the specified time frame. Vertical dotted lines denote mean values.}
    \label{fig: seen_distribution}
\end{figure}

A closer look at the sequence in Figure~\ref{fig: seen_sequence} reveals that there are situations in which vehicles are recognized by FCOs in the past but can no longer be seen in the current time step, e.g., due to occlusion. These vehicles are, therefore, not available per se at the current time to establish a traffic state representation. Let $V_{s,t}$ be the set of vehicles that are either observers or detected in the current or one of the $s$ previous time steps, i.e., $V_{s,t} = \left( \bigcup_{k=t-s}^{t} V_{d,k} \right) \cap V_t$. Since $|V_{s,t}| \geq |V_{d,t}|$, $V_{s,t}$ can generate a richer traffic state representation. Hence, there is temporal enhancing potential in utilizing the detections of previous time steps. Figure~\ref{fig: time_potential_grid} investigates this potential for different penetration rates and sequence lengths for the previously described simulation. For instance, considering a scenario with a 10\% penetration rate and a 20-second sequence length, it is possible to restore approximately 20\% of the vehicles currently present within the intersection, equating to an average of $5.9$ vehicles. When this data is combined with Figure~\ref{fig: seen_distribution}, it can be concluded that the percentage of vehicles available for generating a comprehensive state representation increases significantly, rising from 54\% to 74\% when fully harnessing the temporal potential.

\section{Data-Driven Temporal Enhancement}\label{sec:data-driven-temporal-enhancement}
The preceding section demonstrates that FCOs can identify other traffic participants and construct a traffic state representation. Moreover, the utilization of previous frames of data presents the potential to consider previously detected vehicles. In light of this, we propose a data-driven temporal enhancement of the currently detected vehicles to harness this potential. I.e., we aim for a supervised learning model that learns to recover previously seen vehicles into the current timestep. This also involves inferring the previously seen vehicle's potential change in position and angle into the current timestep.
Training a supervised learning model requires a dataset $D$ that includes information on the set of vehicles $V_{d,t}$ and $V_{s,t}$. Hence, we record dataset entries $D_{d,t} = \{(p_i, \theta_i) \mid i \in V_{d,t}\}$ and $D_{s,t} = \{(p_i, \theta_i) \mid i \in V_{s,t}\}$ for a time window $t_{\text{start}}$ to $t_{\text{end}}$ and step length $\Delta t$ within the simulation. Hereby, $p_i$, $\theta_i$ represent the position and angle of the respective vehicle. 
\newline
During training, the model is provided with a sequence of dataset entries representing consecutive time steps: 
$D_{d,t}, D_{d,t_{-1\Delta t}}, D_{d,t_{-2\Delta t}}, ..., D_{d,t_{-s\Delta t}}$ as input, with $D_{s,t}$ serving as ground truth. 
Instead of utilizing the raw dataset entries as input and target for the model, we propose generating a BEV representation of the vehicles. Thus, employing a BEV representation is similar to the one illustrated in Figure~\ref{fig: intersection} (without building structures). Therefore the model receives a sequence of images, denoted as $S \in {\{0,1\}}^{s \times w \times h}$ as input and ground truth image $G \in {\{0,1\}}^{1 \times w \times h}$, where $s$, $w$, $h$ correspond to the sequence length, width, and height of the image representation, respectively. Such a reformulation of the problem was already successful in trajectory prediction problems, as shown in~\cite{izquierdo2023vehicle}. Employing a BEV representation to train the model offers a significant advantage; it ensures model robustness to the fluctuations in the number of vehicles within the observed scenario and captures spatial-temporal correlations directly within the graphical representation. Constraining the output of the model $O$ within the range of ${[0,1]}^{1 \times w \times h}$ further facilitates a probabilistic interpretation of the results, potentially allowing for an understanding of prediction uncertainties.

\subsection{Model Architectures}\label{subsec: models}

To process the input sequence $S \in {\{0,1\}}^{s \times w \times h}$ and produce an output image $O \in {[0,1]}^{1 \times w \times h}$, various deep learning architectures are being investigated that can capture spatiotemporal relationships. This involves a 3D CNN, which was successfully applied in different video understanding tasks as shown in~\cite{8574597}. A 3D CNN employs convolutions across three dimensions: width, height, and depth (time), effectively processing volumetric data. The network learns spatial-temporal features at each layer by convolving 3D kernels through the input data. As a special case, we will also investigate the effectiveness of a 2D CNN. Here, the third dimension of the CNN kernel matches the input tensor depth. Further, we investigate the Convolutional Long Short-Term Memory (ConvLSTM) model introduced in~\cite{shi2015convolutional}. It combines the special recurrent neural network LSTM that excels in capturing temporal dependencies in sequential data with convolutional layers. Thus allowing to capture of spatial-temporal patterns. 
Inspired by~\cite{arnab2021vivit}, we also investigate an Encoder Temporal-encoder Decoder (ETD) architecture that first creates a one-dimensional embedding using a spatial encoder for each item within the input sequence. I.e. The input sequence $S \in {\{0,1\}}^{s \times w \times h}$ is processed to $S_{\text{emb}} \in \mathbb{R}^{s \times e}$ with $e$ being the chosen embedding size. $S_{\text{emb}}$ is fed into a so-called temporal encoder that extracts the temporal information from the sequence. The temporal encoder thereby generates $O_{\text{emb}} \in \mathbb{R}^{1 \times e}$. Using a decoder to recover the target BEV representation from the embedding and a sigmoid activation function, the final output $O \in {[0,1]}^{1 \times w \times h}$ is generated. Within this structure, we utilize a transformer-based temporal encoder (ETD-Transformer), leveraging self-attention mechanisms to weigh the significance of different parts of the input data, effectively capturing dependencies within the sequential data \cite{vaswani2017attention}. Additionally, we evaluate the LSTM model as the temporal encoder (ETD-LSTM). For the spatial encoder, we employ both a CNN-based and transformer-based Vision Transfomer (ViT) encoder presented in~\cite{dosovitskiy2020image}.

\subsection{Evaluation Metrics}\label{subsec: metrics}
To assess the efficacy of various model architectures, we employ a range of evaluation metrics that effectively capture the ability of the models to retrieve previously detected vehicles in the current time step accurately. First, we use the Intersection Over Union (IOU), which quantitatively measures the overlap between a predicted segmentation and the ground truth. Similar to~\cite{rößle2024unlocking}, we transform the output $O\in {[0,1]}^{1 \times w \times h}$ into a binary mask $O_{\text{bin}} \in {\{0,1\}}^{1 \times w \times h}$ using a threshold $\tau$, which we set to $0.5$ in this work. $O_{\text{bin}}$ can thus be interpreted as a segmentation of areas of the BEV representation occupied by vehicles. Following this, the IoU can be calculated as: 
\begin{equation}
\text{IoU} = \frac{|O_{\text{bin}} \cap G|}{|O_{\text{bin}} \cup G|}, 
\label{eq: iou}
\end{equation} 
allowing a direct comparison of occupied areas by vehicles. 
With the IoU itself, it is not possible to determine the number of vehicles recovered from past time steps and positioned correctly at the current time step. We therefore investigate the number of Recovered Vehicles (RV). Given the output of the model $O_{\text{bin}}$, we formally calculate them as: 
\begin{equation}
\text{RV} = \sum_{i=1}^{|V_{s,t}|} \mathbf{1}\left( (v_{i} \notin V_{d,t}) \land \left(\frac{|v_{i,\text{BEV}} \cap O_{\text{bin}}|}{|v_{i,\text{BEV}}|} \geq \theta\right)\right).
\label{eq: rv} 
\end{equation}
Here, $v_{i,\text{BEV}}$ represent the portion of the BEV representation that is occupied by $v_i \in V_{s,t}$, $\theta$, set to $0.5$ in this work, expresses the threshold necessary for the vehicle to be recovered and $\mathbf{1}$ represents the indicator function. Following this, we introduce the Rnhidden Vehicle Metric (RVM), which sets the RV in relation to the temporal enhancement potential, i.e., the difference between $V_{s,t}$ and $V_{d,t}$. 
\newline
As vehicles that are actually available in $V_d,t$ can also be lost through the processing of the model, we also examine the Lost Vehicles (LV), which can formally be described as: 
\begin{equation}
\text{LV} = \sum_{i=1}^{|V_{t,d}|} \mathbf{1} \left(\frac{|v_{i,\text{BEV}} \cap O_{\text{bin}}|}{|v_{i,\text{BEV}}|} < \theta\right).
\label{eq: uvm}
\end{equation}
\newline
Again, we set this value relative to the temporal enhancement potential to achieve a Lost Vehicle Metric (LVM).
Additionally, it is important to account for instances where the model erroneously infers the presence of vehicles in $O$
 not existing in $V_{s,t}$ or inferring them in the wrong location, effectively hallucinating vehicles. To quantitatively assess this, we introduce a Hallucinated Pixel Metric (HPM), defined as:  
 \begin{equation}
\text{HPM} = \frac{1}{\sum_{i,j} \mathbf{1}(G_{ij} = 1)}\sum_{i,j} \mathbf{1} (O_{ij,\text{bin}} - G_{ij}).
\label{eq: hpm}
\end{equation}
Hence, the HPM effectively captures the false positive predictions on a pixel level. The LVM and UVM accommodate on a vehicle level and capture the true positive (UVM) and false negative (LVM) predictions. The IoU encapsulates the other metrics and additionally captures the shape of the reconstructed vehicles.

\begin{figure}[t]
    \centering
        \includegraphics[width=\columnwidth]{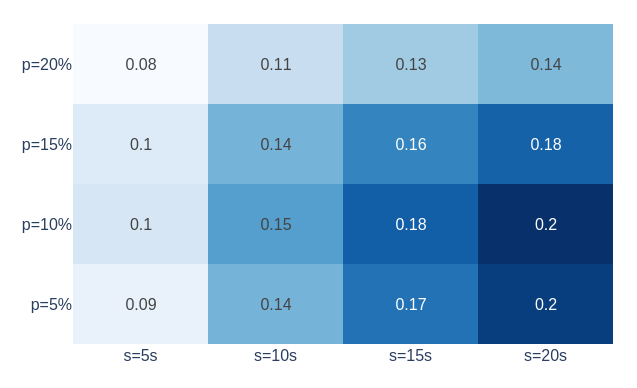}
    \caption{Mean difference in $V_{s,t}$ and $V_{d,t}$ relative to $V_t$ across various FCO penetration rates and sequence lengths showing the temporal enhancing potential.}
    \label{fig: time_potential_grid}
\end{figure}
\subsection{Training Methodology}
For training various model architectures aimed at enhancing temporal data, inspired by~\cite{xu2022cobevt}, we employ a weighted Binary Cross-Entropy Loss ($\mathcal{L}_{\text{BCE}}$), setting $w_{\text{BCE}}$ to $2$. This choice of $w_{\text{BCE}}$ is designed to counterbalance the disparity between the presence of vehicles and background in the BEV representations. 
The dataset was created as described in Section \ref{sec:data-driven-temporal-enhancement} for the intersection previously analyzed in Section \ref{subsec: FCO}. To enhance the dataset size and ensure a proper division into training, validation, and testing subsets, we executed simulations between $t_{\text{start}}$~= 6~AM to $t_{\text{start}}$~=~9~AM with $\Delta t$~=~1s on ten separate occasions. Each instance involved a variation in the initial traffic demand provided by~\cite{harth2021automated}. Specifically, we introduced variability in the vehicles' entry times to the traffic network by applying a normal distribution with a standard deviation of 15 seconds, hence generating different traffic scenarios at the investigated intersection. For the size of the BEV representation, we employ $512 \times 512$ pixels that cover a radius of 100m around the intersection. Eight of those runs with different random seeds are used for training, one for validation, and one for testing.
The 3D CNN, 2D CNN, and ConvLSTM are trained for 25 epochs with the training dataset. We employ a pre-training strategy for the ETD structures that omit the temporal-encoder component. Consequently, the architecture undergoes pre-training in an autoencoder fashion, compressing the BEV representations into embedding space, $O_{\text{emb}} \in \mathbb{R}^{1 \times e}$, where we employ $e=1024$ as the dimensionality of the embeddings. After pre-training 25 epochs, the complete architectures using the LSTM and transformer-based temporal encoder are again trained for 25 epochs. 

\subsection{Results}
\input{images/multi_images/input_sequence}
\input{images/multi_images/qualitative_results_new}
\input{tables/results}
The initial pre-training of the ETD structures shows that the CNN-based spatial encoder outperforms the transformer-based encoder and achieves an IoU of $0.94$ for the test set with an LVM of 0. Thus, the encoder can compress the BEV representation into an embedding space of $e=1024$ without losing valuable information. 
An exemplary input sequence for the full training of the spatiotemporal models is shown in Figure 
\ref{fig: input_sequence}. Figure \ref{img: qualitative_results} visualizes the ground truth and the qualitative results for the different models using the input using that input sequence. Notably, the 3D CNN architecture demonstrates that is able to recover vehicles that are not moving within the sequence. However, the architecture loses the vehicles currently in motion, even though they are detectable in the last timestep. In contrast, the 2D CNN and ConvLSTM also successfully recovered stationary vehicles without suffering from lost vehicles. Remarkably, the ETD architectures emerge as the only models proficient in accurately reconstructing the movement trajectories of vehicles into the current frame. Thus, it shows the capability of not only recovering stationary vehicles but also vehicles in motion at their current position. However, it is pertinent to acknowledge that these advanced architectures do not guarantee the detection of all vehicles. Challenges arise, particularly when vehicles are only visible in a limited number of frames within $S$. Comparing the ETD-LSTM and ETD-Transformer, it can be noted that the transformer-based encoder positions the moving vehicles more accurately than the LSTM leading, leading to a lower HVM. Those qualitative results also reflect the quantitative results evaluating the average metrics for the test dataset, which are presented in Table \ref{tab: results}. Thus, LSTM and transformer architectures achieve the highest UVM with 41\% and 36\% for a sequence length of five seconds, whereas the transformer-based architecture outperforms the LSTM in lower LVM and HVM. The 2D CNN architecture achieves the highest average IOU since it successfully recovers stationary vehicles with an LVM of zero and perfectly reconstructs the vehicles' shape in the BEV representation. 
Table \ref{tab: results} further illustrates that the temporal enhancement remains effective for longer sequence durations. For sequences spanning 20 seconds, the ETD-LSTM network successfully identifies 40\% of the vehicles that are currently not observable. This is particularly noteworthy as longer sequences exhibit increased potential for vehicle detection, as detailed in Figure \ref{fig: time_potential_grid}. In absolute terms, this translates to an average recovery of 2.4 vehicles, which are additionally made available for creating a traffic state representation.

\section{Conclusion}
In this work, we analyzed the potential of using FCOs for generating a traffic state representation, which is crucial for different management strategies. We achieved this by emulating the 3D detectability of traffic participants given the camera sensors of the FCO. The analysis conducted at a specific intersection revealed that even modest penetration rates of FCOs can provide significant information about the other vehicles within the simulation, thereby demonstrating the potential of FCOs to replace traditional stationary observation methods. Further analysis revealed that additional information can be obtained by correlating detections from past timesteps with the current timesteps, i.e., showcasing a temporal enhancement potential. We proposed a data-driven methodology that utilizes spatiotemporal deep-learning architectures to leverage this capability. This approach aims to re-identify previously seen vehicles in the current timestep. Results reveal that up to 41\% of the vehicles seen in the sequence of the previous timestep but undetectable (e.g., due to occlusions) in the current time step can be successfully recovered. The findings additionally demonstrate stability in longer sequences, where in absolute numbers, more vehicles can be recovered. For future work, we propose the integration of physical driving models into the data-driven temporal data enhancement process to facilitate the recovery of vehicles from the previous time steps. Finally, the traffic state information generated by the FCOs needs to be investigated to be integrated into traffic management algorithms and compared to other traffic monitoring systems.

\section*{Acknowledgment}
This work was partially funded by the Bavarian state government as part of the High Tech Agenda and BayWISS.


\end{document}

%% file: images/multi_images/detected_sequence.tex
\begin{figure*}[tb]
      \begin{minipage}{\textwidth}
        \centering
        \includegraphics[width=0.98\textwidth]{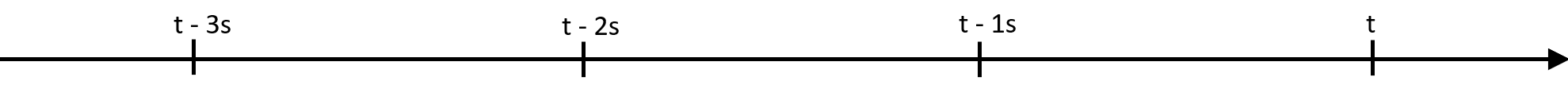}
    \end{minipage}
\begin{minipage}[t]{0.245\textwidth}
    \centering
    \adjustbox{width=\textwidth, height=5cm, keepaspectratio, trim=15mm 1mm 0 0, clip}{%
        \includegraphics{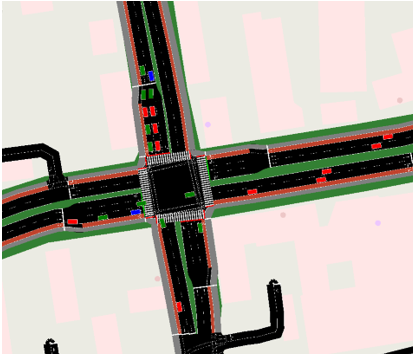}
    }
\end{minipage}%
\hfill 
\begin{minipage}[t]{0.245\textwidth}
    \centering
    \adjustbox{width=\textwidth, height=5cm, keepaspectratio, trim=15mm 0 0 0, clip}{%
        \includegraphics{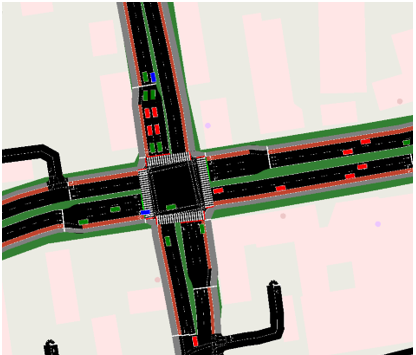}
    }
\end{minipage}%
\hfill 
\begin{minipage}[t]{0.245\textwidth}
    \centering
    \adjustbox{width=\textwidth, height=5cm, keepaspectratio, trim=15mm 1mm 0 0, clip}{%
        \includegraphics{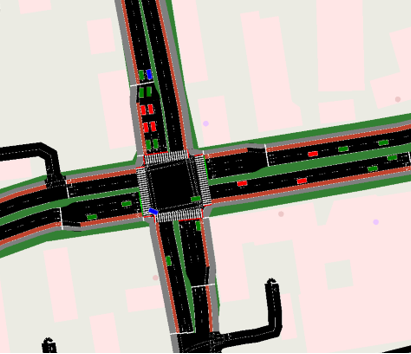}
    }
\end{minipage}%
\hfill 
\begin{minipage}[t]{0.245\textwidth}
    \centering
    \adjustbox{width=\textwidth, height=5cm, keepaspectratio, trim=15mm 0 0 0, clip}{%
        \includegraphics{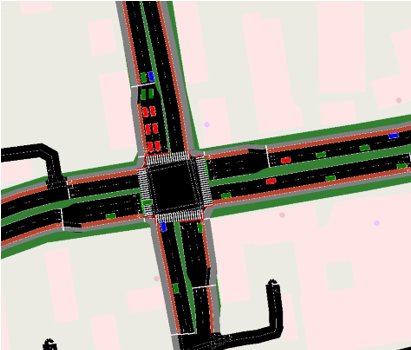}
    }
\end{minipage}
    \caption{Sequential visualization of FCOs (blue), detected vehicles (green), and undetected vehicles (red) across scenarios at one-second intervals. The BEV emulation approach is utilized to determine the 3D detectability of vehicles.} 
    \label{fig: seen_sequence}
\end{figure*}

%% file: images/multi_images/input_sequence.tex
\begin{figure*}[b]
    \centering
    \begin{minipage}{\textwidth}
        \centering
        \includegraphics[width=\textwidth]{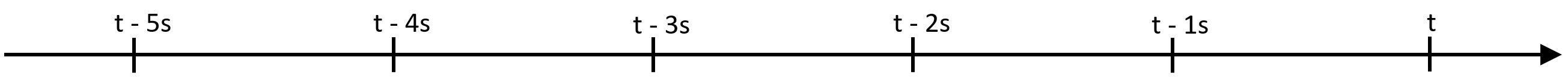}
    \end{minipage}
    \vspace{0cm}
    \begin{minipage}[b]{0.16\textwidth}
        \centering
        \includegraphics[width=\linewidth]{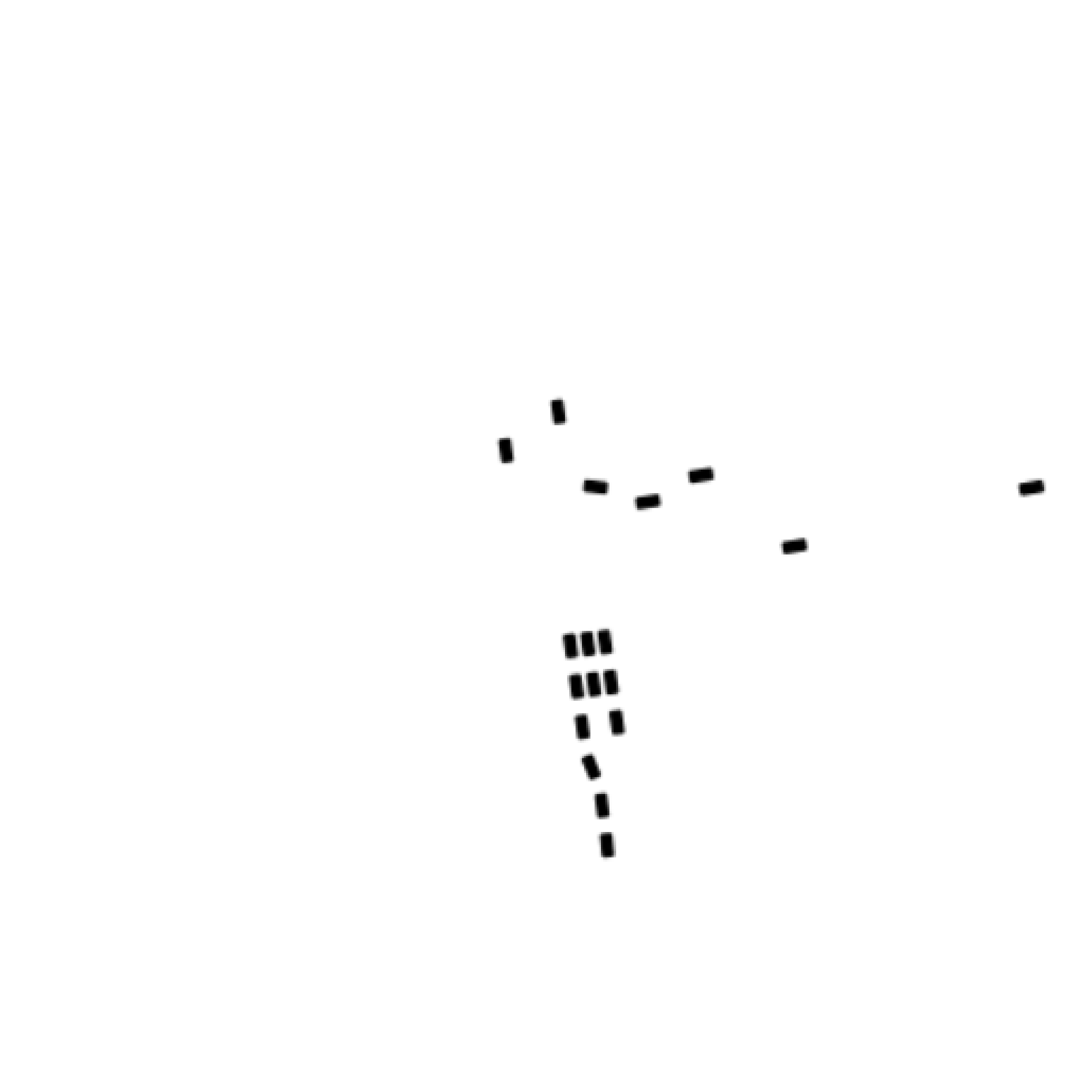}
    \end{minipage}%
    \hfill
    \begin{minipage}[b]{0.16\textwidth}
        \centering
        \includegraphics[width=\linewidth]{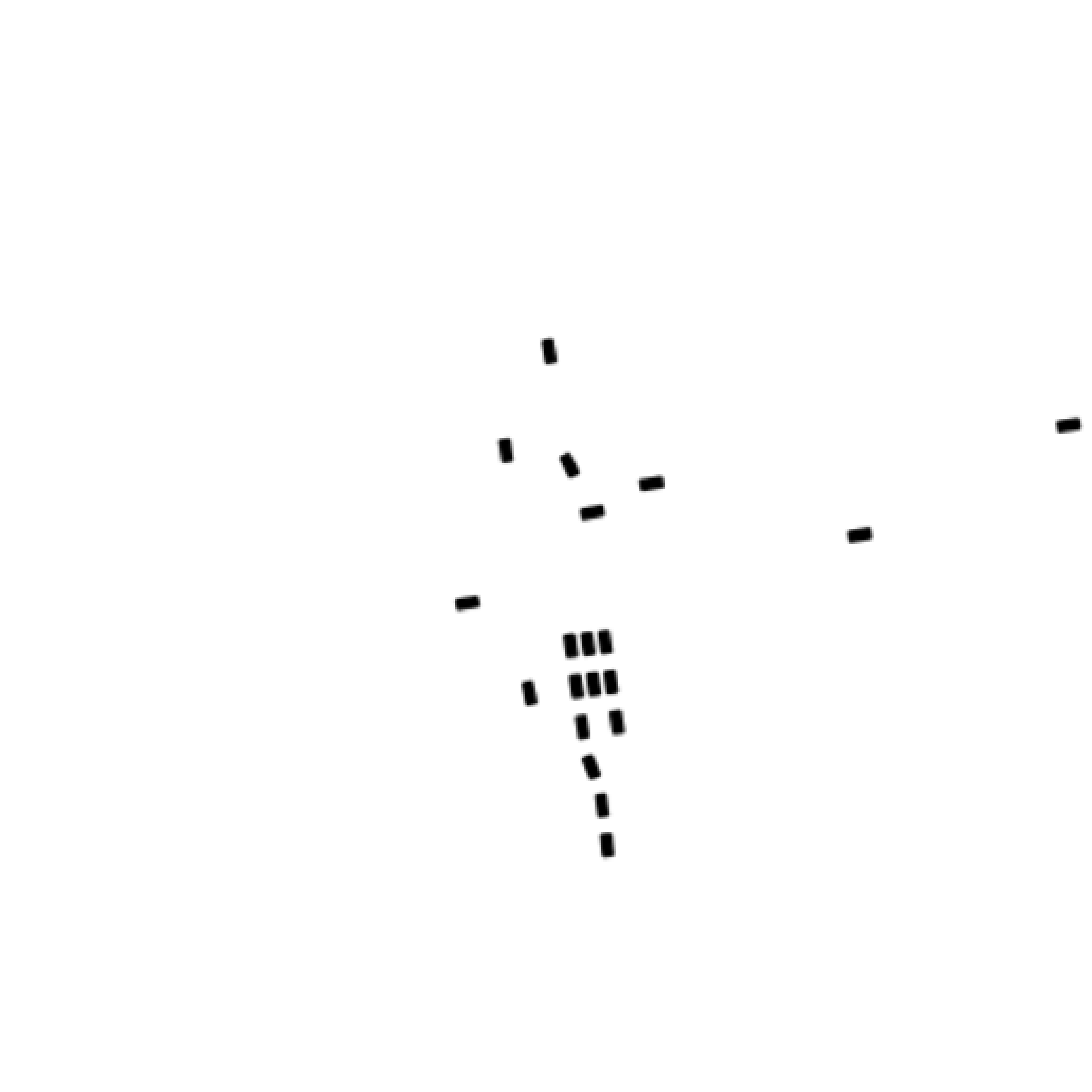}
    \end{minipage}%
    \hfill
    \begin{minipage}[b]{0.16\textwidth}
        \centering
        \includegraphics[width=\linewidth]{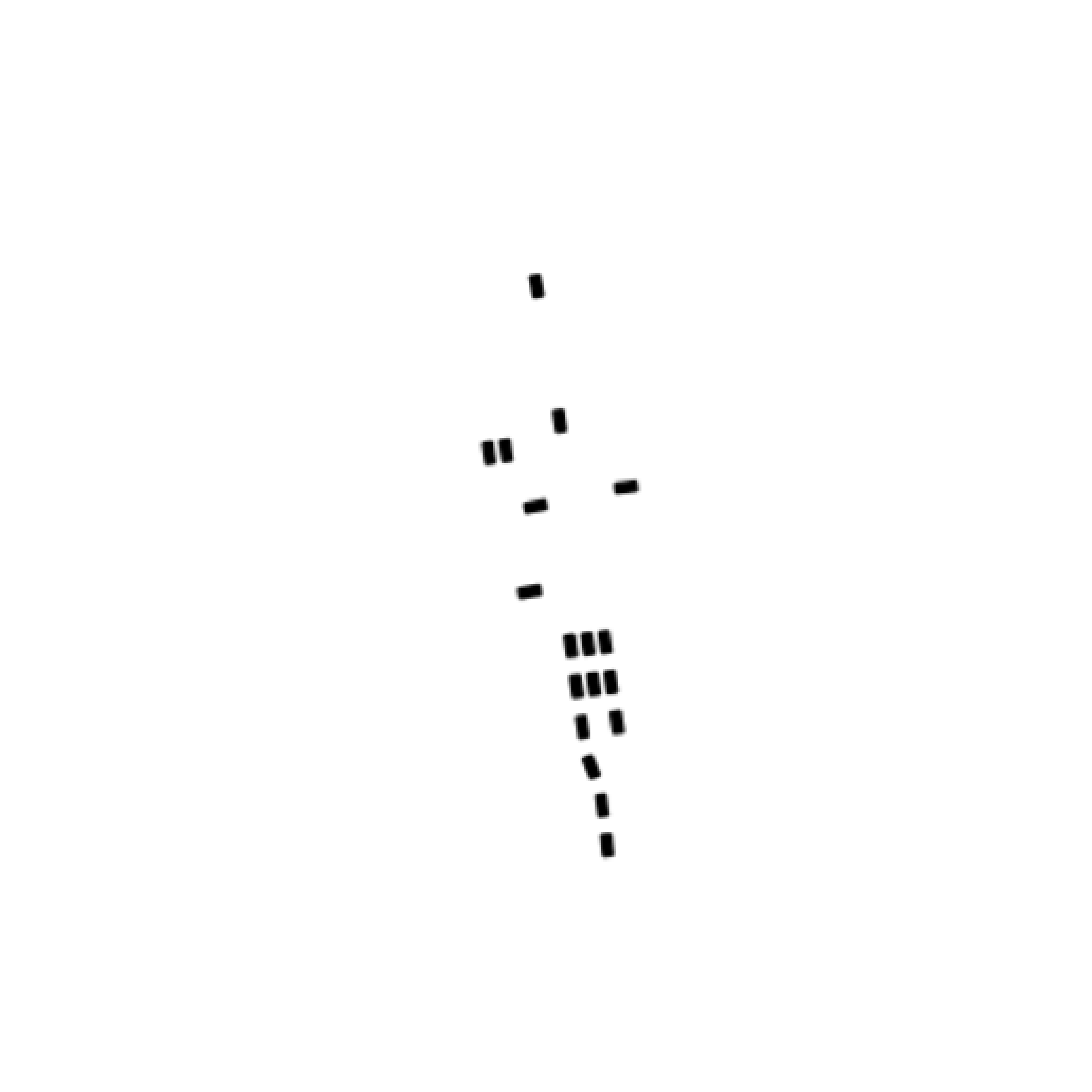}
    \end{minipage}%
    \hfill
    \begin{minipage}[b]{0.16\textwidth}
        \centering
        \includegraphics[width=\linewidth]{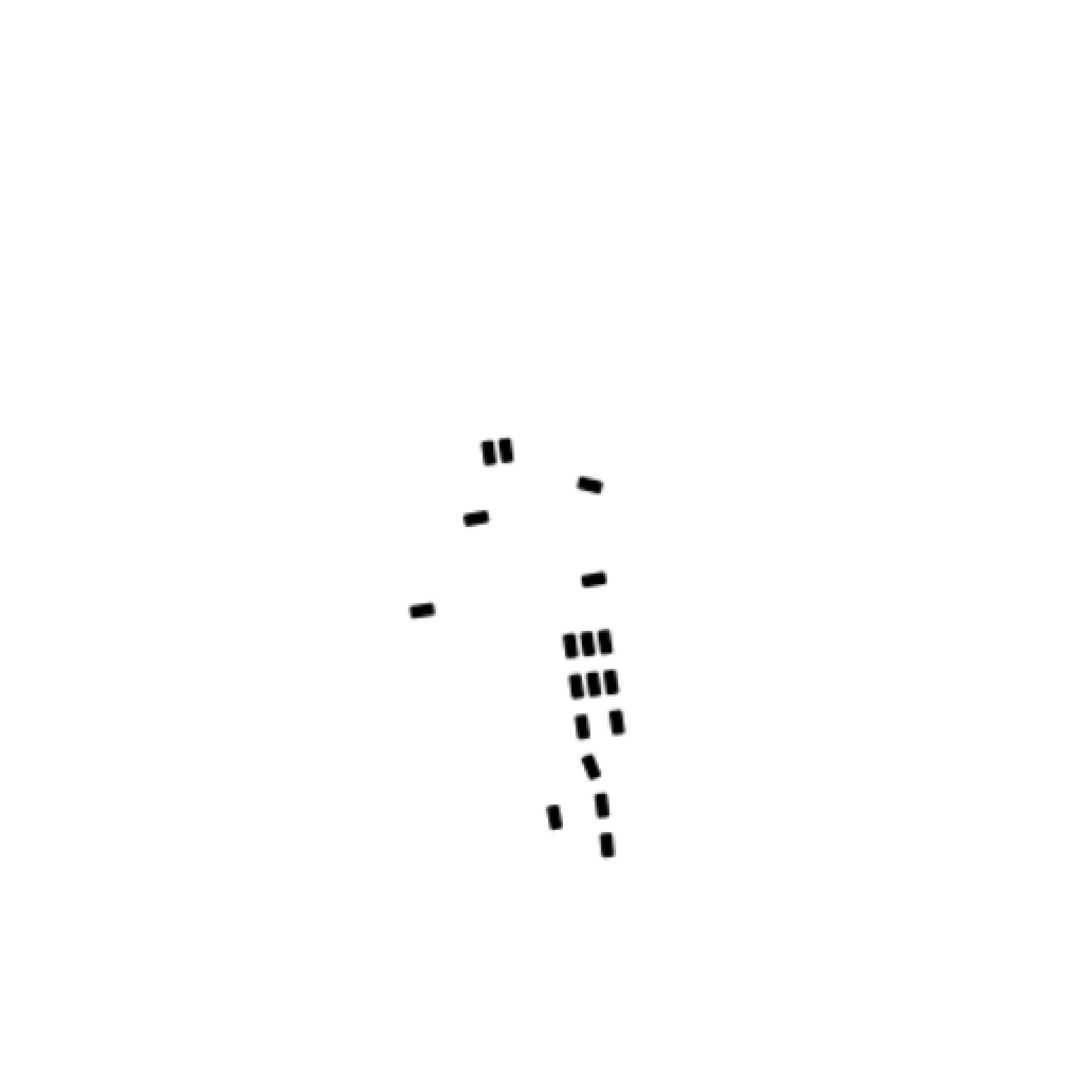}
    \end{minipage}%
    \hfill
    \begin{minipage}[b]{0.16\textwidth}
        \centering
        \includegraphics[width=\linewidth]{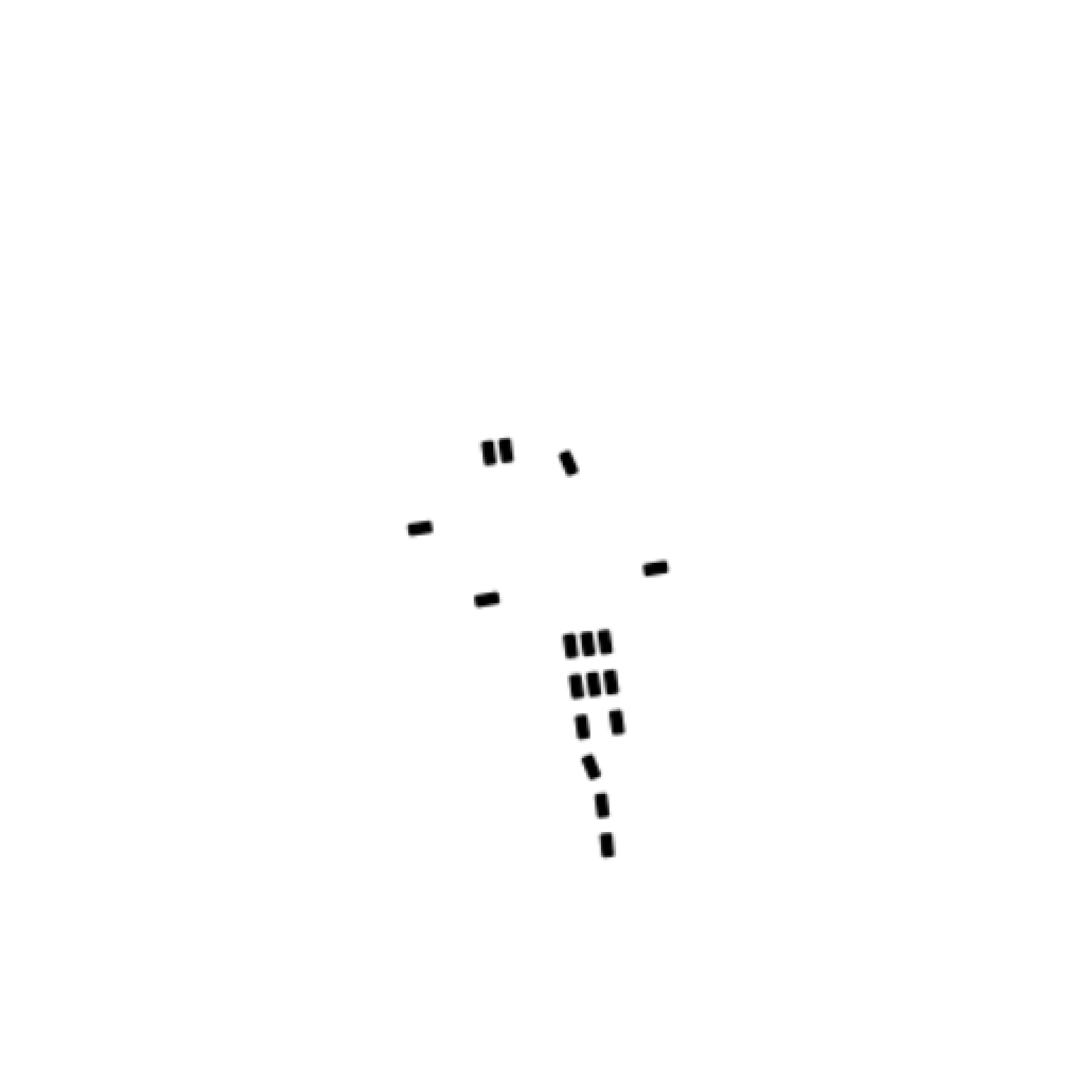}
    \end{minipage}%
    \hfill
    \begin{minipage}[b]{0.16\textwidth}
        \centering
        \includegraphics[width=\linewidth]{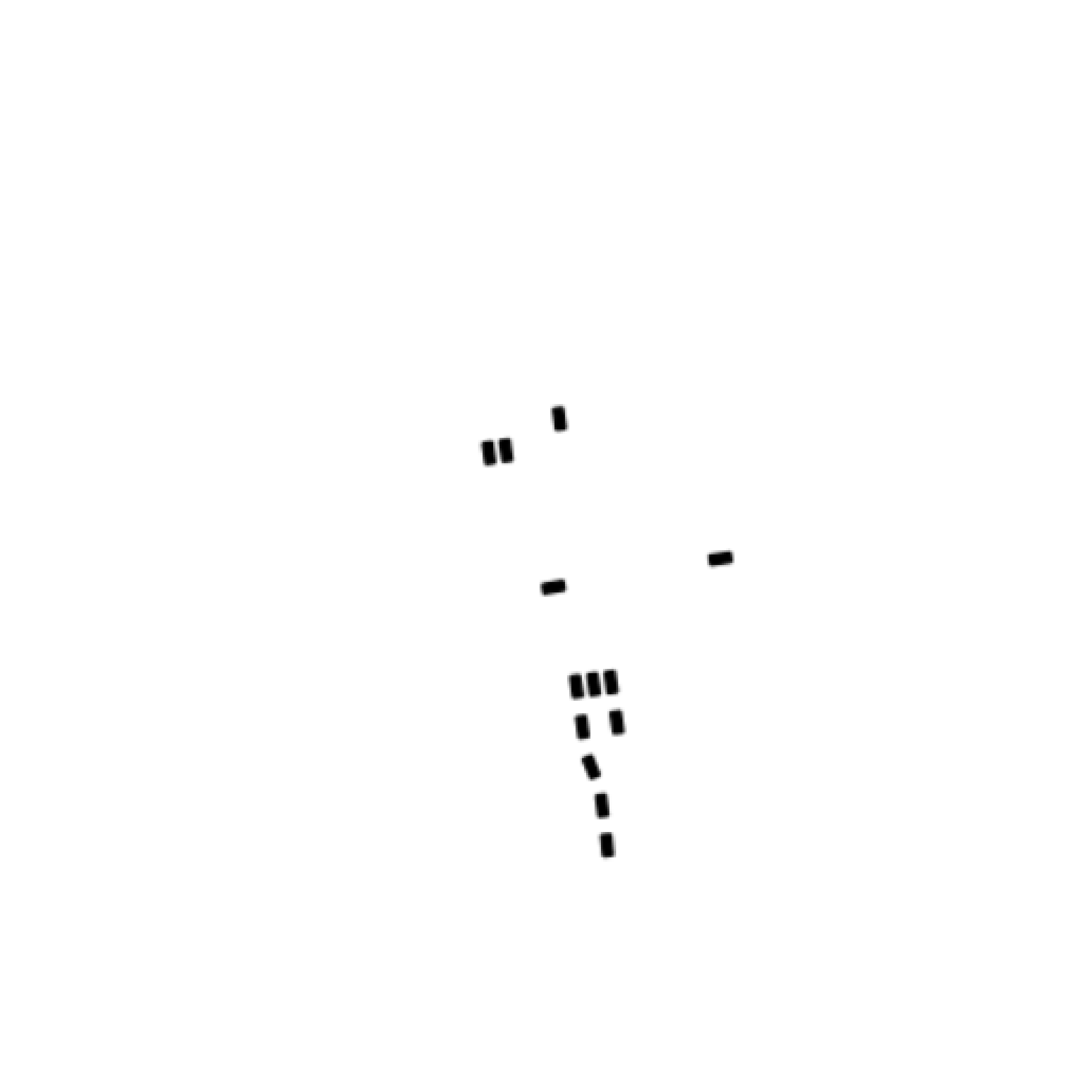}
    \end{minipage}%
    \caption{Examplary sequence $S \in {\{0,1\}}^{6 \times 512 \times 512}$ showing $V_{d,t-5s}$ to $V_{d,t}$ in BEV representation. The sequence serves as the input to the spatiotemporal models.}
    \label{fig: input_sequence}
\end{figure*}

%% file: images/multi_images/qualitative_results_new.tex
\begin{figure*}[b]
    \centering
    \begin{minipage}[b]{0.16\textwidth}
        \centering
        \includegraphics[width=\linewidth]{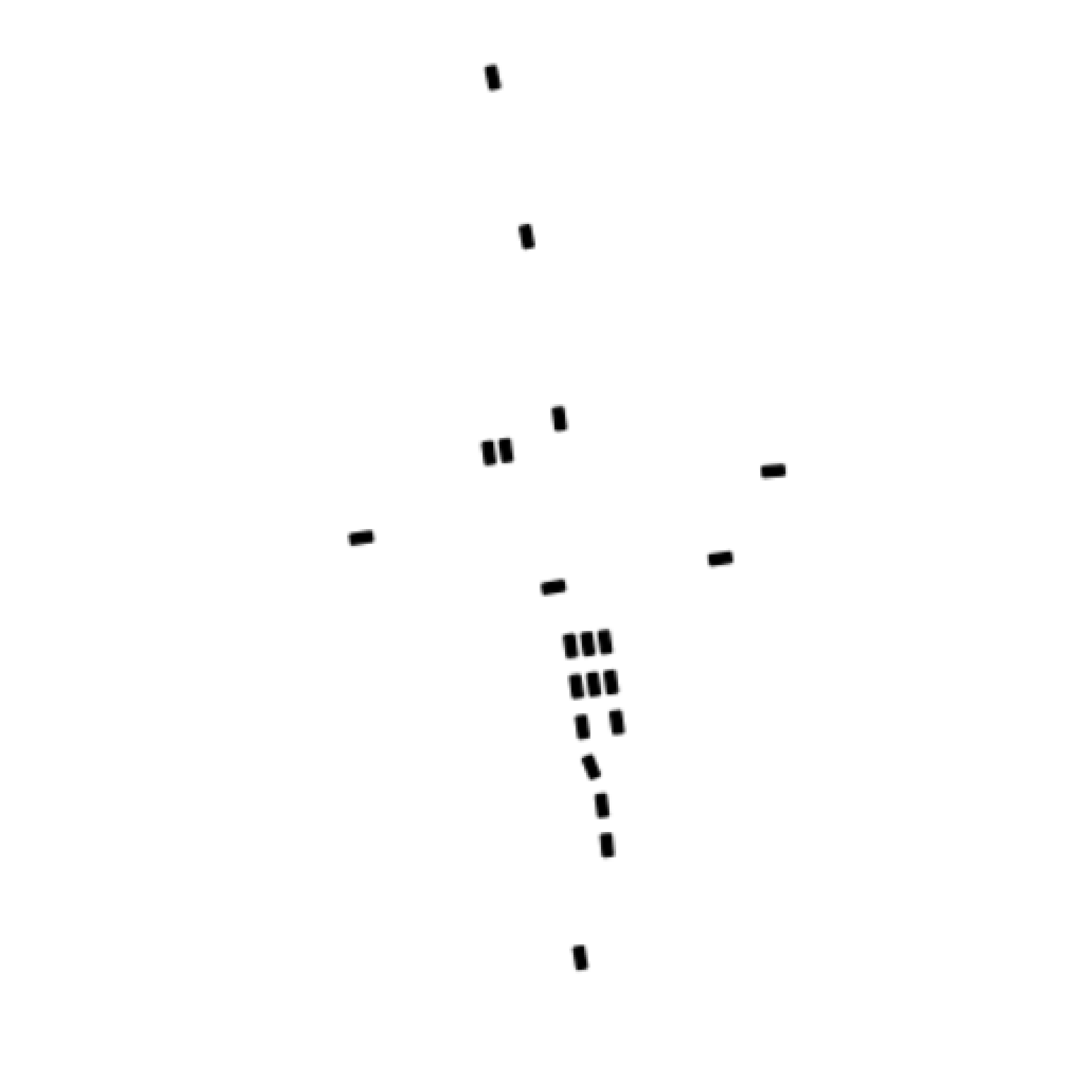}
    \end{minipage}%
    \hfill
    \begin{minipage}[b]{0.16\textwidth}
        \centering
        \includegraphics[width=\linewidth]{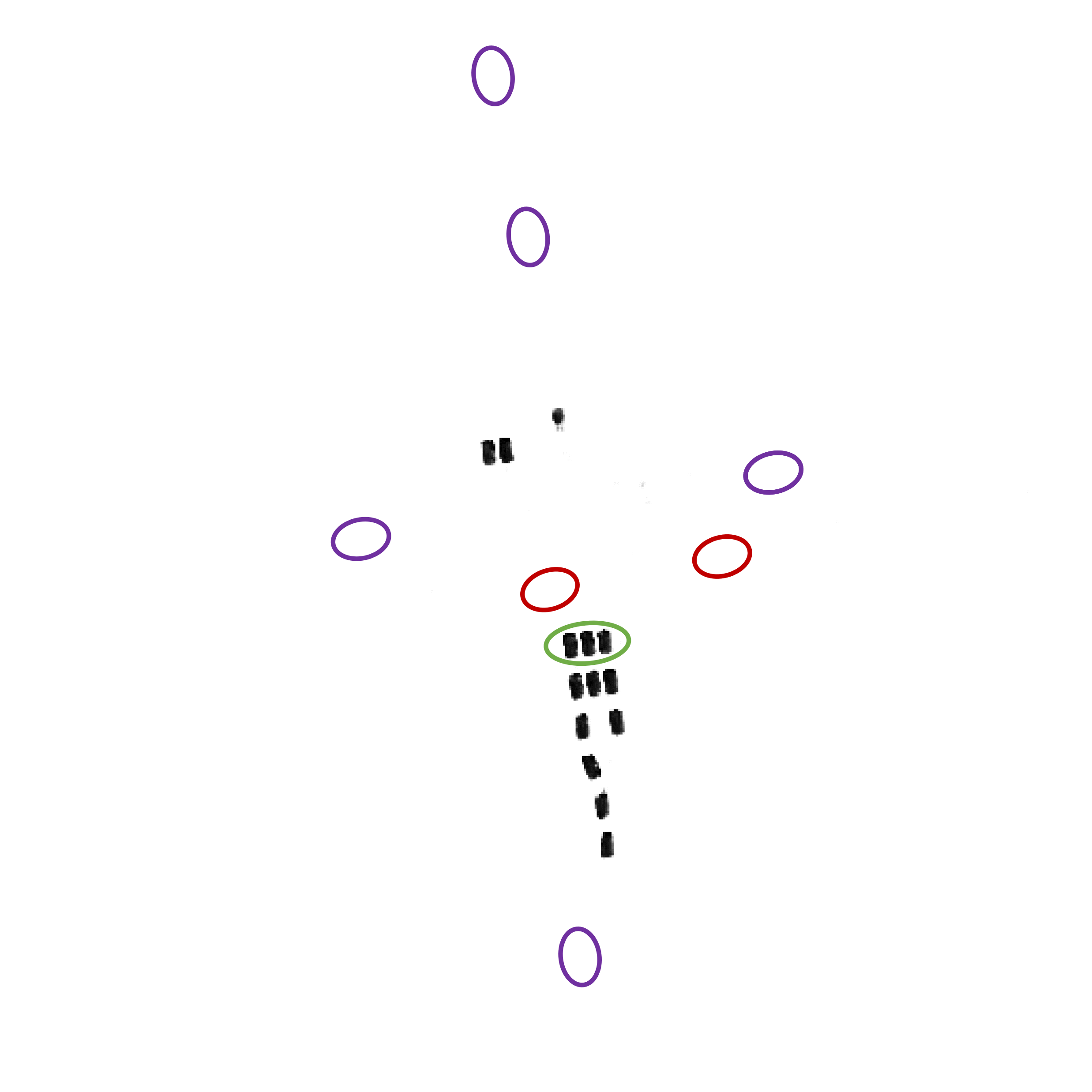}
    \end{minipage}%
    \hfill
    \begin{minipage}[b]{0.16\textwidth}
        \centering
        \includegraphics[width=\linewidth]{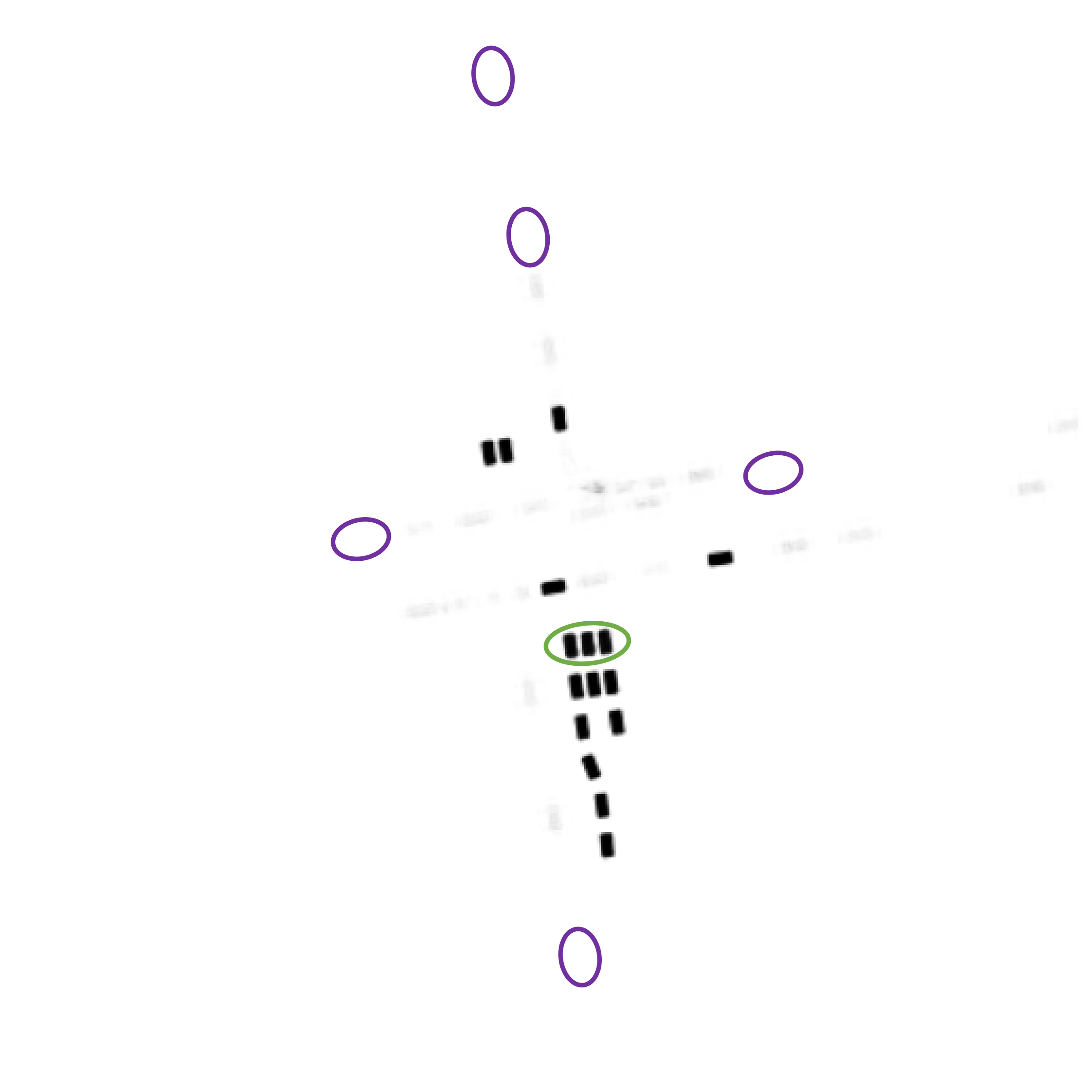}
    \end{minipage}%
    \hfill
    \begin{minipage}[b]{0.16\textwidth}
        \centering
        \includegraphics[width=\linewidth]{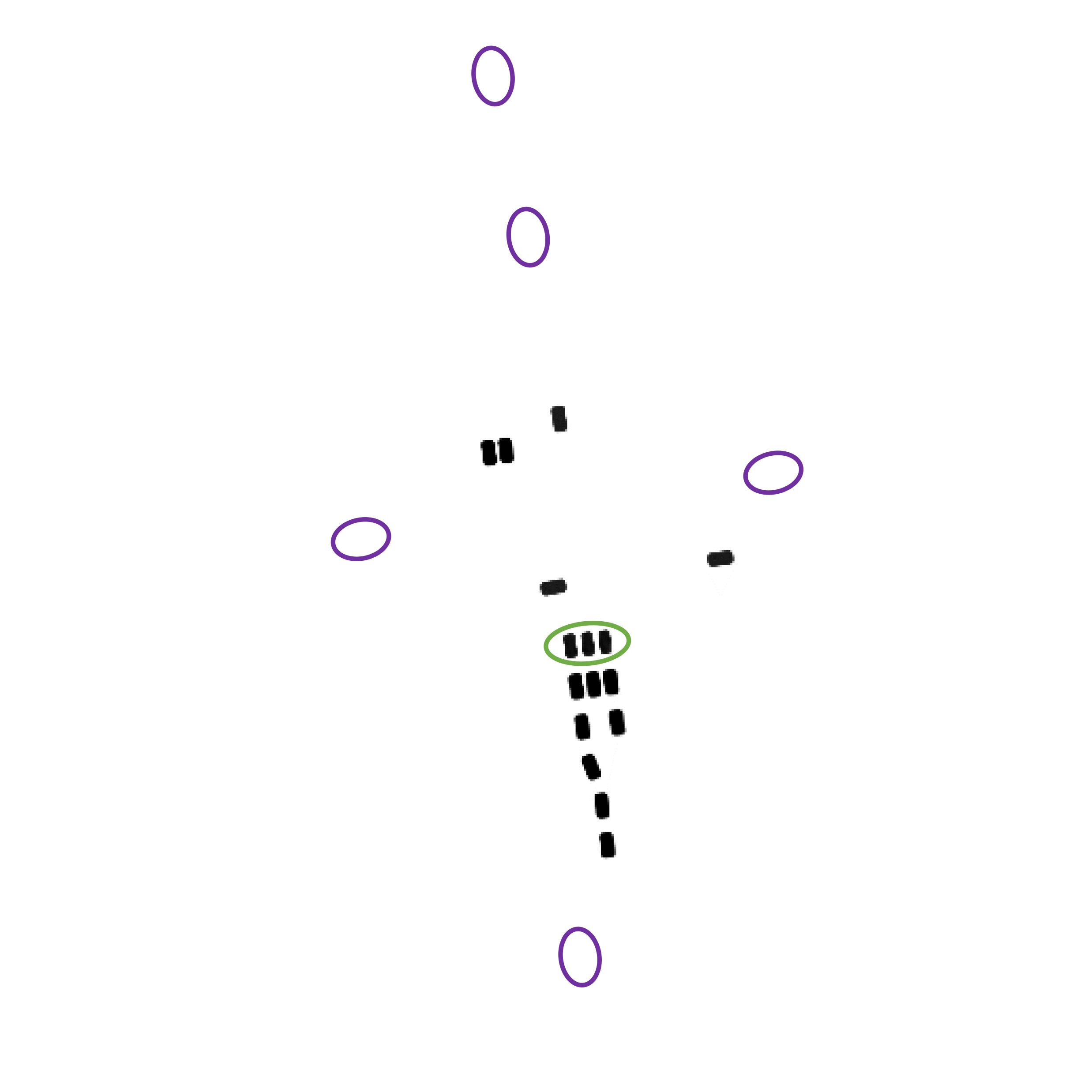}
    \end{minipage}%
    \hfill
    \begin{minipage}[b]{0.16\textwidth}
        \centering
        \includegraphics[width=\linewidth]{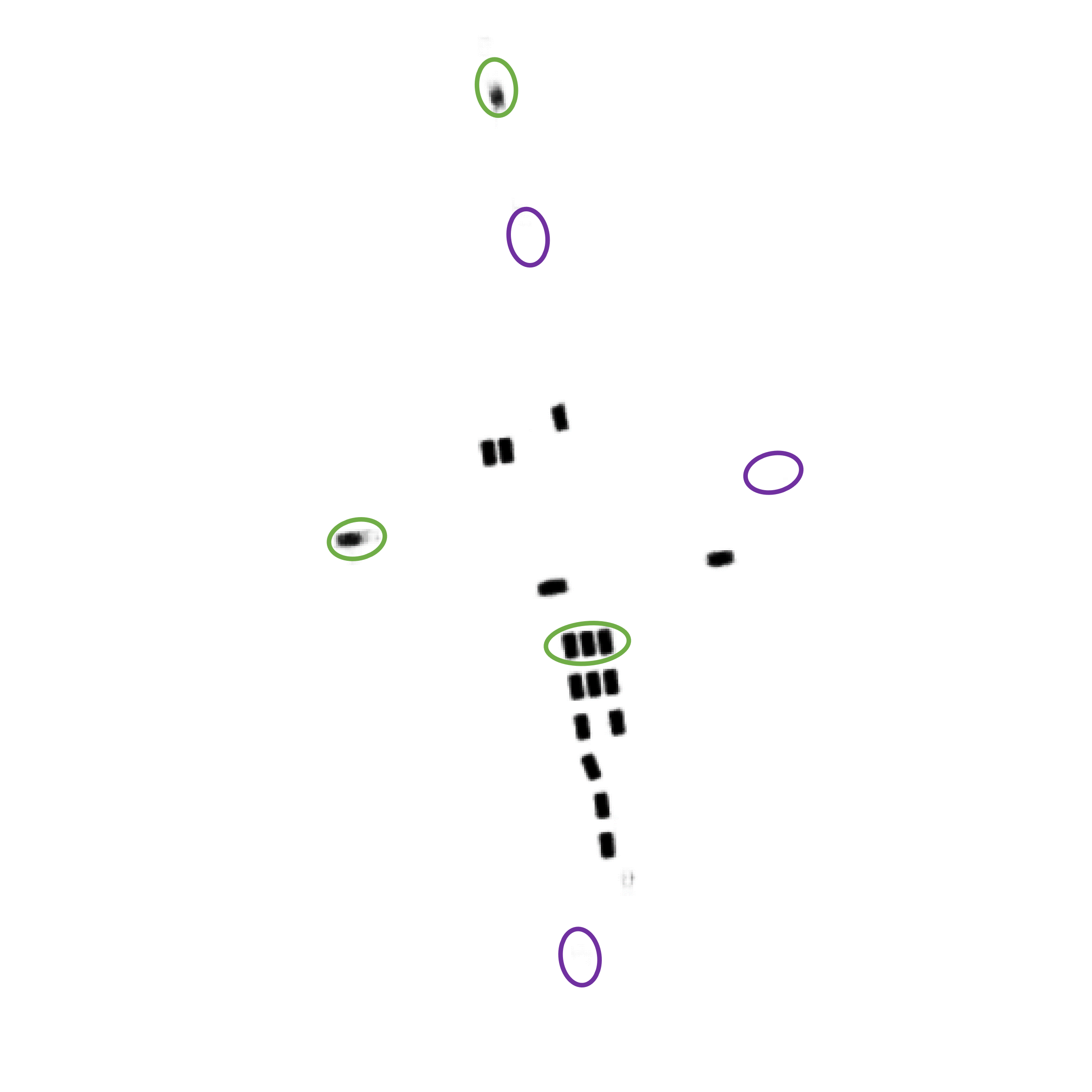}
    \end{minipage}%
    \hfill
    \begin{minipage}[b]{0.16\textwidth}
        \centering
        \includegraphics[width=\linewidth]{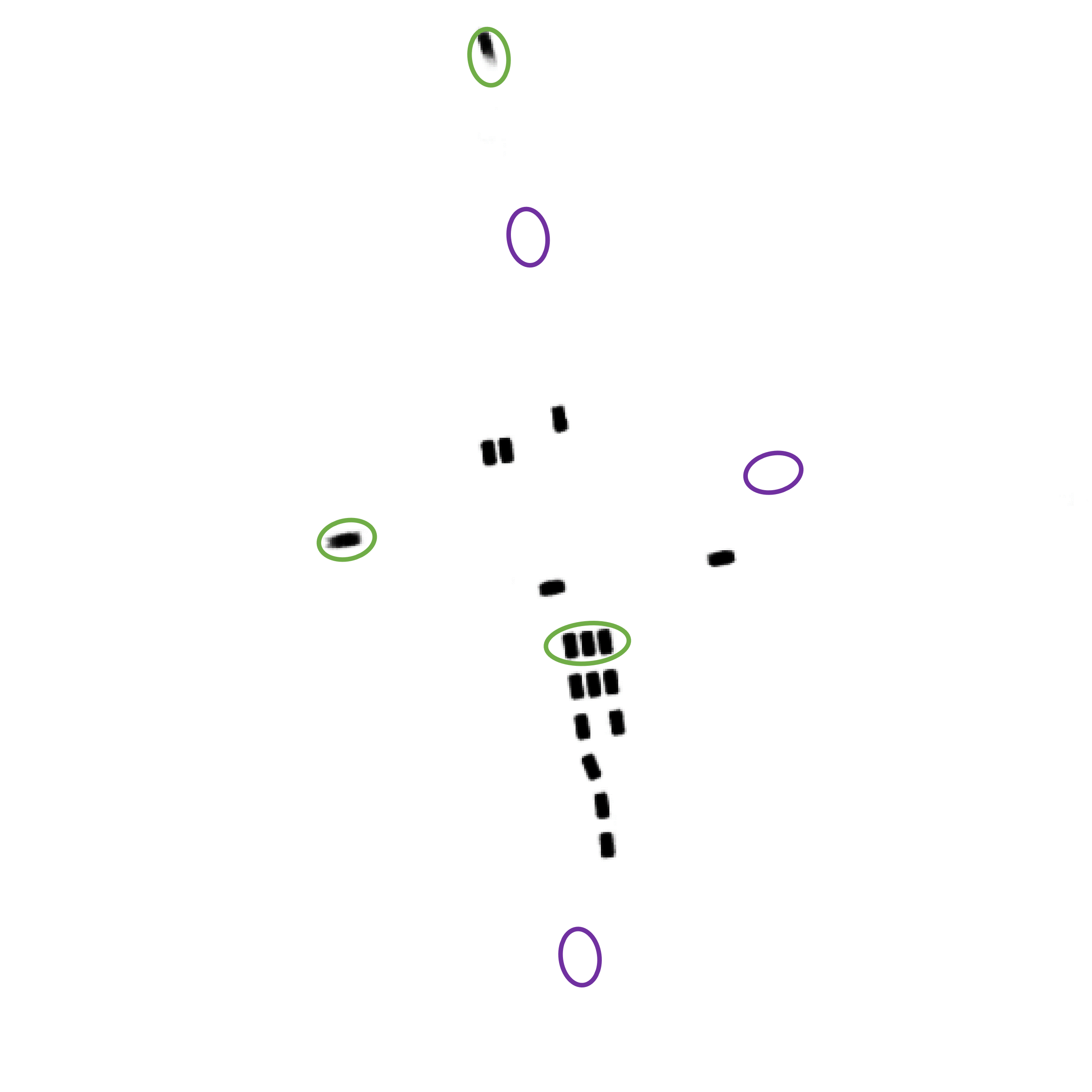}
    \end{minipage}%
    
    \begin{minipage}[b]{0.16\textwidth}
        \centering
        \caption*{Ground Truth}
    \end{minipage}
    \hfill
    \begin{minipage}[b]{0.16\textwidth}
        \centering
        \caption*{3D CNN}
    \end{minipage}
    \hfill
        \begin{minipage}[b]{0.16\textwidth}
        \centering
        \caption*{2D CNN}
    \end{minipage}
    \hfill
        \begin{minipage}[b]{0.16\textwidth}
        \centering
        \caption*{CONV-LSTM}
    \end{minipage}
    \hfill
        \begin{minipage}[b]{0.16\textwidth}
        \centering
        \caption*{ETD-LSTM}
    \end{minipage}
    \hfill
        \begin{minipage}[b]{0.16\textwidth}
        \centering
        \caption*{ETD-Transformer}
    \end{minipage}
    \hfill

    \caption{Ground truth $G \in {\{0,1\}}^{1 \times 512 \times 512}$ for the sequence $S$ shown in Figure \ref{fig: input_sequence} and the outputs $O\in{[0,1]}^{1 \times 512 \times 512}$ for the different models. The outputs are added with highlights for the lost vehicles (red), not recovered vehicles (violet), and successfully recovered vehicles (green).}
    \label{img: qualitative_results}
\end{figure*}


%% file: tables/results.tex
\begin{table}[tb]
    \centering
    \caption{Quantitative evaluation of model performance for sequence lengths of 5 and 20 seconds, featuring various models outlined in Section~\ref{subsec: models} and assessed using metrics introduced in Section~\ref{subsec: metrics}. Results show mean values for the test dataset.}
    \label{tab: results}
    \begin{tabular}{c|cccc}
        \hline
        Evaluation Metric & IoU & RVM & LVM & HVM \\ \hline \hline
        \multicolumn{5}{c}{$s=5s$} \\ \hline
        3D CNN & $0.48$ & $0.30$ & $1.77$ & $0$\\ 
        2D CNN & $0.79$ & $0.27$ & $0$ & $0.01$ \\ 
        Conv LSTM & $0.67$ & $0.21$ & $0$ & $0.23$ \\ 
        ETD-LSTM & $0.70$ & $0.41$ & $0.09$ & $0.18$ \\ 
        ETD-Transformer & $0.73$ & $0.36$ & $0.02$ & $0.13$ \\ 
        \hline
        \multicolumn{5}{c}{$s=20s$} \\ \hline
        3D CNN & $0.38$ & $0.25$ & $1.80$ & $0$\\
        2D CNN & $0.63$ & $0.27$ & $0$ & $0.17$ \\ 
        Conv LSTM & $0.56$ & $0.21$ & $0$ & $0.23$ \\
        ETD-LSTM & $0.64$ & $0.40$ & $0.07$ & $0.24$ \\ 
        ETD-Transformer & $0.68$ & $0.35$ & $0.01$ & $0.20$ \\ 
        \hline
        \hline
    \end{tabular}
\end{table}